\documentclass[letterpaper,journal]{IEEEtran}

\usepackage{amsmath,amsfonts,amssymb,bm}
\usepackage{algorithmic}
\usepackage{algorithm}
\usepackage{array}
\usepackage[caption=false,font=normalsize,labelfont=sf,textfont=sf]{subfig}
\usepackage{textcomp}
\usepackage{stfloats}
\usepackage{url}
\usepackage{verbatim}
\usepackage{graphicx}
\usepackage{cite}
\usepackage{booktabs}
\usepackage{multirow}
\usepackage{xcolor}
\usepackage[hidelinks]{hyperref}

\title{SCMA: Structure-Conditioned and Metal-Aware Flow Matching for CT Metal Artifact Reduction}

\author{Heran Wang, Jianing Sun, Xu Jiang, Genwei Ma, Jigang Duan, and Xing Zhao%
\thanks{This work was supported by the National Natural Science Foundation of China (Grant No. 12426308), the Beijing High Innovation Plan ("Capital High-End Leading Talents Aggregation and Cultivation Program", Grant No. 202504841094), and the National Key Research and Development Program of China (Grant No. 2020YFA0712200).
(Corresponding authors: Jigang Duan (e-mail: hiduanjigang@163.com).) and Xing Zhao (e-mail: zhaoxing\_1999@126.com) }%
\thanks{Heran Wang, Xu Jiang, Jigang Duan, and Xing Zhao are with the School of Mathematical Sciences, Capital Normal University, Beijing 100048, China.}%
\thanks{Jianing Sun is with the College of Mathematics Science, Inner Mongolia Normal University, Hohhot 010022, China.}%
\thanks{Genwei Ma is with the National Center for Applied Mathematics Beijing, Capital Normal University, and the Academy for Multidisciplinary Studies, Capital Normal University, Beijing 100048, China.}}

\begin{document}

\maketitle

\begin{abstract}
In X-ray computed tomography (CT), strong attenuation by metallic objects causes beam hardening, photon starvation, and scattering, making measured projections deviate from the ideal imaging model and producing streaks, dark bands, and structural distortions that compromise clinical diagnosis and quantitative analysis. Existing metal artifact reduction (MAR) methods remain limited: optimization-based methods may leave residual artifacts or blur structures, regression networks may generalize poorly across scenarios, and generative models without sample-specific structural guidance and physical constraints may produce anatomically inconsistent structures. Flow Matching learns a continuous-time velocity field that deterministically transports a source distribution to a target distribution, providing a flexible prior for MAR. However, standard unconditional Flow Matching does not exploit sample-specific structure, the spatially nonuniform nature of metal-induced degradation, or original projection measurements. To address these limitations, we propose SCMA, a structure-conditioned and metal-aware Flow Matching framework. First, a linear-interpolation-corrected image is jointly fed into the velocity network with the intermediate state as a sample-specific structural condition, guiding inference toward artifact-free CT images while preserving anatomy. Second, time-varying spatial weights derived from the metal mask and its distance transform are incorporated into the Flow Matching loss to emphasize severe degradation within and around metal regions. Finally, conditional Flow Matching updates alternate with projection-consistency correction during inference, allowing reliable measurements outside metal traces to constrain predictions. Experiments on simulated and real CT data demonstrate that SCMA more effectively suppresses metal artifacts, preserves local anatomical structures, and reduces hallucination-like structures inconsistent with projection measurements than representative MAR methods.
\end{abstract}


\section{Introduction}

\IEEEPARstart{C}{omputed} tomography (CT) has been widely used in medical diagnosis, radiotherapy planning, industrial nondestructive testing, and security inspection because of its noninvasiveness, high spatial resolution, and strong structural depiction capability. When metallic objects are present in the scanned object, X-rays passing through the metal undergo severe attenuation, leading to beam hardening, photon starvation, and scattering. These effects cause projection measurements along metal-intersecting paths to deviate from the ideal imaging model, producing metal artifacts such as streaks, dark bands, and local structural distortions in reconstructed images. Metal artifacts not only reduce image readability and quantitative accuracy but also affect downstream tasks such as segmentation and registration. Therefore, effectively suppressing metal artifacts while preserving genuine structures remains an important problem in CT imaging~\cite{selles2024advances,withers2021xray}.

\subsection{Existing Methods and Limitations}

Traditional metal artifact reduction (MAR) methods mainly rely on projection completion, physical modeling, and regularized reconstruction. Projection completion methods, represented by linear interpolation (LI)~\cite{kalender1987reduction} and normalized metal artifact reduction (NMAR)~\cite{meyer2010normalized}, typically identify metal traces in the projection domain from image-domain metal regions and then restore the corrupted projections through interpolation~\cite{prell2009forward}, normalization-based completion~\cite{meyer2012fsmar}, or frequency decomposition~\cite{anhaus2022nonlinearly,anhaus2023nonlinear}. Physical modeling methods reduce model mismatch by explicitly accounting for polychromatic X-ray attenuation, beam hardening, and related effects~\cite{verburg2012beam,park2016beam}. Regularized reconstruction methods improve reconstruction stability using iterative correction~\cite{wang1996iterative}, total variation minimization~\cite{zhang2011constrained}, superiorized iteration~\cite{humphries2020superiorized}, region-adaptive regularization~\cite{wei2019regional}, or sparse priors~\cite{mehranian2013sparsity}. Although traditional MAR methods offer good physical interpretability, their performance is limited by the accuracy of metal-trace estimation, physical modeling, and manually designed priors. When the metal objects are large, numerous, or surrounded by complex structures, these methods may still produce residual artifacts, secondary artifacts, and structural blurring.

Recent deep learning methods learn nonlinear relationships between metal-corrupted data and artifact-free CT images in a data-driven manner. According to the processing domain, existing methods can be categorized as image-domain, projection-domain, and dual- or multi-domain approaches. Image-domain methods typically take artifact-affected images or conventional MAR results as input and directly predict corrected images~\cite{zhang2018cnn,liao2020adn,wang2022dicdnet,wang2024oscnet}. Projection-domain methods mainly complete or correct corrupted data within the metal traces~\cite{ghani2019deepmar,yu2020dscmar,yu2021selfsupervised}. Dual- and multi-domain methods further combine image-domain structural information with projection-domain measurements to improve artifact suppression and structure preservation~\cite{zhou2022dudodr,wang2023indudonet,du2023udamar,li2024quad,shi2024coupling,su2024f2iflow,wang2025dual}.

From the perspective of modeling paradigms, existing learning-based MAR methods mainly employ regression and generative models. Regression-based methods typically learn a deterministic mapping from corrupted data to artifact-free images. However, many of these methods rely on synthetically paired data, and the discrepancy between synthetic degradation and real measurements may limit their generalizability across scanners, acquisition protocols, and real-world scenarios~\cite{li2026robust,ma2026universal}. Generative methods provide stronger image priors by modeling the distribution of high-quality CT images or conditional restoration distributions. Generative adversarial networks~\cite{lee2021betacyclegan}, diffusion models~\cite{karageorgos2024ddpm,liu2024diffusionpriors,cai2024diffmar,luo2025bcddmar,yang2025ctsdm}, score-based generative models~\cite{wu2024waveletscore,zhang2024waveletmultichannel}, and other continuous generative models~\cite{xu2024sword,wu2024multichannel,wu2026universal} have increasingly been applied to MAR and related CT image restoration tasks. However, when sample-specific conditions are insufficient or measurement constraints are absent from the generative process, these models may produce visually plausible structures that are inconsistent with the anatomy of the current sample or the original projection measurements. Therefore, generative MAR should exploit not only image-distribution priors but also sample-specific structural information and reliable projection consistency.

Flow Matching (FM) is a continuous normalizing-flow-based generative modeling method that learns a time-dependent velocity field to continuously transport samples from a source distribution to a target distribution through a deterministic ordinary differential equation~\cite{lipman2022flow}. Rather than learning a one-step mapping from degraded images to target images, FM learns the local transport direction of each intermediate state along a continuous probability path, providing a direct training objective, an explicit generative trajectory, and a controllable inference process. However, standard unconditional FM provides only a population-level prior over the target image distribution. The transport from a Gaussian distribution to artifact-free CT images neither describes metal-induced degradation nor incorporates the structure, metal location, or original projection measurements of the current sample. Therefore, a continuous generative path alone does not make FM a restoration model tailored to MAR. Applying FM to CT metal artifact reduction requires sample-specific structural conditions, metal-region priors, and CT projection measurement constraints.

\subsection{Motivation}

\begin{figure}[t]
    \centering
    \includegraphics[width=\linewidth]{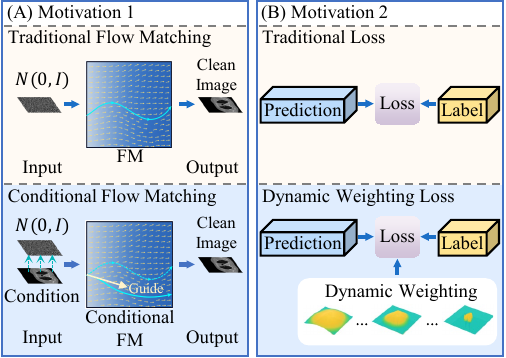}
    \caption{Two Flow Matching designs tailored to MAR in SCMA. (A) LI-guided conditional Flow Matching uses an LI image to provide a coarse structural condition for the current sample. (B) Metal-region dynamic weighting constructs a time-varying two-dimensional pixel-wise weight map using the metal mask and distance information.}
    \label{fig:motivation}
\end{figure}

The objective of CT metal artifact reduction is not merely to generate visually plausible artifact-free images, but to recover the genuine structure of the current sample in severely degraded regions while maintaining consistency with reliable projection measurements. Based on this objective, we adapt Flow Matching to MAR from three perspectives: sample-specific structural conditioning, metal-region learning, and projection measurement constraints.

First, as shown in Fig.~\ref{fig:motivation}(A), standard FM starts from a Gaussian sample $\boldsymbol{z}\sim\mathcal{N}(\boldsymbol{0},\boldsymbol{I})$, where $\boldsymbol{I}$ denotes the identity matrix, and transports it toward the artifact-free image distribution along the learned velocity field. Without a condition derived from the current sample, the model must infer the complete structure solely from the population-level image prior, introducing uncertainty into the restoration. Although the LI image still contains interpolation errors, residual artifacts, and structural blurring and therefore cannot serve as the final corrected result, it retains the coarse structure of the current sample. We thus jointly feed the intermediate state $\boldsymbol{x}_{t}$ and the LI image $\boldsymbol{x}_{\mathrm{Linear}}$ into the velocity network at each time point, as indicated by the orange arrow. In this way, the FM trajectory is transformed from unconditional population-level generation into a conditional restoration process constrained by the structure of the current sample.

Second, metal artifacts are spatially nonuniform. Severe degradation is typically concentrated within and around the metal regions, whereas areas farther from the metal are relatively reliable. A spatially uniform FM loss can be dominated by large areas with mild degradation, weakening the model's ability to learn the severe artifacts near the metal. As shown in Fig.~\ref{fig:motivation}(B), we construct a two-dimensional pixel-wise weight map $\boldsymbol{W}_{t}$ at each time point using the metal mask $\boldsymbol{M}$ and its distance-transform map $\boldsymbol{D}$, and apply it to the velocity prediction error. During the early stage near the noise endpoint, the weighting covers a broader neighborhood around the metal to enhance the learning of large-scale streaks and dark bands. As the state approaches the artifact-free image endpoint, the weighted region gradually contracts, allowing the model to focus on the metal regions and their immediate neighborhoods while reducing disturbances to reliable structures farther away.

Finally, relying solely on an image-domain generative prior may still produce hallucination-like structures inconsistent with the CT measurements. We therefore introduce projection-consistency correction (PCC) at each inference stage. PCC uses the original metal-affected projection $\boldsymbol{y}_{\mathrm{MA}}$, the metal trace $\boldsymbol{T}$, and the forward-projection operator $\mathcal{F}(\cdot)$ to constrain the current prediction with reliable observations outside the metal traces, and feeds the correction residual back into subsequent Flow Matching updates. By combining LI-based structural conditioning, metal-region dynamic spatial weighting, and projection-consistency correction, SCMA imposes task-specific constraints on sample structure, degraded regions, and physical measurements, rather than directly applying generic Flow Matching to CT image generation.

\subsection{Our Contribution}

Based on the above designs, the main contributions of this work are summarized as follows:
\begin{itemize}
    \item We propose SCMA, a structure-conditioned and metal-aware Flow Matching framework for CT metal artifact reduction. SCMA introduces an LI-corrected image into velocity-field learning as a sample-specific structural condition, enabling the generative trajectory to restore artifact-free CT images under the coarse structural constraint of the current sample and thereby reducing the structural uncertainty associated with unconditional generation.
    \item We propose a dynamically spatially weighted Flow Matching loss guided by the metal mask. The proposed loss constructs time-varying spatial weights from the metal mask and its distance transform, enabling the model to adaptively focus on the metal regions and their neighborhoods at different generative stages and improving the restoration of locally severe degradation.
    \item We develop an alternating inference framework that integrates the Flow Matching generative prior with projection-consistency correction. The same velocity network is shared across all time steps, while reliable projections outside the metal traces constrain the current prediction. This design requires no separate network for each inference time step, reduces hallucination-like structures inconsistent with the original measurements, and improves the physical reliability of the MAR results.
\end{itemize}

\section{Related Work}
\label{sec:related_work}

\subsection{Flow Matching Generative Models}
\label{subsec:related_flow_matching}

Flow Matching (FM) is a class of generative modeling methods based on continuous normalizing flows. It learns a time-dependent velocity field that continuously transports samples from a source distribution to a target data distribution through a deterministic ordinary differential equation~\cite{lipman2022flow}. Let $\boldsymbol{x}_{1}$ denote a sample drawn from the target data distribution and $\boldsymbol{z}\sim\mathcal{N}(\boldsymbol{0},\boldsymbol{I})$ denote Gaussian noise, where $\boldsymbol{I}$ is the identity matrix. FM constructs a linear probability path between them as
\begin{equation}
\boldsymbol{x}_{t}
=
(1-t)\boldsymbol{z}
+
t\boldsymbol{x}_{1},
\qquad
t\in[0,1],
\label{eq:related_fm_path}
\end{equation}
where $t=0$ corresponds to the noise endpoint and $t=1$ to the data endpoint. The target velocity associated with this path is
\begin{equation}
\boldsymbol{u}_{t}
=
\frac{d\boldsymbol{x}_{t}}{dt}
=
\boldsymbol{x}_{1}
-
\boldsymbol{z}.
\label{eq:related_fm_velocity}
\end{equation}
The velocity network $\boldsymbol{v}_{\theta}(\boldsymbol{x}_{t},t)$ is trained by regressing the target velocity, with the basic objective defined as
\begin{equation}
\mathcal{L}_{\mathrm{FM}}
=
\mathbb{E}_{\boldsymbol{x}_{1},\boldsymbol{z},t}
\left[
\left\|
\boldsymbol{v}_{\theta}
\left(
\boldsymbol{x}_{t},t
\right)
-
\boldsymbol{u}_{t}
\right\|_{2}^{2}
\right].
\label{eq:related_fm_loss}
\end{equation}
Once trained, samples from the target distribution can be obtained by starting from $\boldsymbol{x}_{0}=\boldsymbol{z}$ and solving
\begin{equation}
\frac{d\boldsymbol{x}_{t}}{dt}
=
\boldsymbol{v}_{\theta}
\left(
\boldsymbol{x}_{t},t
\right).
\label{eq:related_fm_ode}
\end{equation}
FM directly learns the local transport directions along a probability path and therefore provides an explicit training objective, a continuous inference trajectory, and flexible numerical solvers.

However, standard unconditional FM learns only the population-level distribution of target images and does not establish a correspondence between a random initial sample and a specific metal-affected image. Its velocity field incorporates neither the structural condition and metal-region information of the current sample nor constraints from the original CT projections. Consequently, unconditional FM can provide only a distributional prior over artifact-free CT images and cannot directly serve as a sample-specific MAR model. Building on this formulation, we further introduce LI-based structural conditioning, dynamic spatial weighting over metal regions, and projection-consistency correction, such that the FM inference trajectory is jointly constrained by the structure of the current sample, the spatial distribution of artifacts, and reliable projection measurements.

\subsection{Metal Segmentation in MAR}
\label{subsec:related_metal_segmentation}

Metal-region segmentation is a preprocessing step in many MAR methods. Its results are commonly used to locate metal regions in the image domain and determine the corresponding metal traces through forward projection. Traditional methods mainly employ CT-number thresholding and morphological processing to identify metal regions, but they are prone to undersegmentation or oversegmentation in the presence of severe streak artifacts, high-density bone structures, or ambiguous boundaries. To improve localization robustness, Hegazy et al. used U-Net to directly predict metal traces from projection data~\cite{hegazy2019unet}. U-Net employs an encoder--decoder architecture with skip connections to integrate multiscale semantic information and spatial details~\cite{ronneberger2015unet}. nnU-Net further uses a self-configuring mechanism to automatically determine the preprocessing procedure, network architecture, training strategy, and postprocessing pipeline, demonstrating strong adaptability across various medical image segmentation tasks~\cite{isensee2021nnunet}.

Metal segmentation serves different purposes in different MAR modules. In projection completion methods such as LI and NMAR, the segmentation result directly determines the extent of the metal traces to be replaced. Undersegmentation may retain corrupted measurements, whereas oversegmentation may discard reliable projections. These methods are therefore generally sensitive to metal-boundary localization. In this work, we employ an existing nnU-Net to obtain the image-domain metal mask $\boldsymbol{M}$ rather than developing a new segmentation network. During training, $\boldsymbol{M}$ and its distance information are used to adjust the loss weights at different spatial locations. During preprocessing and inference, $\boldsymbol{M}$ is forward-projected to obtain the metal trace $\boldsymbol{T}$, which is used to generate the LI-corrected conditioning image and delineate the reliable measurement regions for projection-consistency correction. In loss weighting, the mask serves as a spatial attention prior, whereas the projection-domain operations remain affected by the accuracy of metal-trace localization. We therefore regard metal segmentation as an auxiliary preprocessing step rather than a methodological contribution of SCMA.

\subsection{Linear Interpolation-Based MAR}
\label{subsec:related_li_mar}

Linear interpolation-based metal artifact reduction (LI-MAR) is a classical projection completion method~\cite{kalender1987reduction,prell2009forward}. It first generates projection-domain metal traces from image-domain metal regions and regards measurements within these traces as unreliable. Linear interpolation is then performed using the nonmetal measurements on both sides of each metal trace at the same projection angle. The completed projections are subsequently reconstructed to obtain an LI-corrected image. LI-MAR is simple to implement and computationally efficient, and it can effectively reduce severe streak and dark-band artifacts. However, it essentially replaces missing or distorted measurements within the metal traces with manually interpolated values. When the metal traces are wide or the projections on their two sides differ substantially, the interpolated values cannot accurately represent the true projection variation, potentially introducing secondary artifacts, intensity bias, and structural blurring.

Normalized metal artifact reduction (NMAR) uses the forward projection of a prior image to normalize the original projections, performs interpolation in the normalized domain, and subsequently restores the projection magnitudes, thereby reducing the influence of anatomical variations on interpolation~\cite{meyer2010normalized,meyer2012fsmar}. Nevertheless, the performance of NMAR still depends on the quality of the prior image and the accuracy of metal-trace localization. Unlike methods that directly use the LI result as the final output, we employ $\boldsymbol{x}_{\mathrm{Linear}}$ only as a coarse, sample-specific structural condition. Although it still contains residual artifacts and local structural errors, its preserved principal anatomical contours can constrain the Flow Matching inference trajectory, while subsequent restoration is jointly performed by the conditional velocity field and projection-consistency correction.

\section{Method}
\label{sec:method}

\begin{figure*}[!t]
    \centering
    \includegraphics[width=\textwidth]{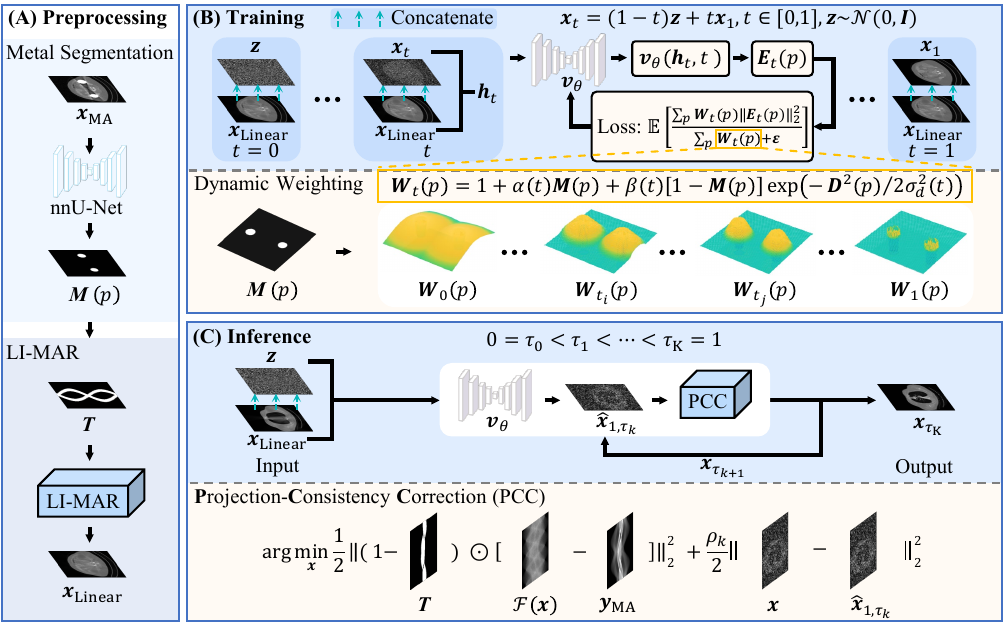}
    \caption{Overall workflow of SCMA. (A) In the preprocessing stage, metal segmentation is performed to obtain the image-domain metal mask, and LI-MAR is applied to generate the conditioning image $\boldsymbol{x}_{\mathrm{Linear}}$. (B) During training, intermediate states are densely sampled along the Flow Matching path, with $\boldsymbol{x}_{\mathrm{Linear}}$ used as the conditioning input and the metal mask used to construct a dynamically weighted loss. (C) During inference, the process starts from Gaussian noise and evolves under the fixed condition $\boldsymbol{x}_{\mathrm{Linear}}$, with projection-consistency correction (PCC) applied at each stage.}
    \label{fig:method_overview}
\end{figure*}

We propose SCMA, a structure-conditioned and metal-aware Flow Matching framework for CT metal artifact reduction. As illustrated in Fig.~\ref{fig:method_overview}, SCMA consists of three stages: preprocessing, conditional velocity-field training, and inference with projection-consistency correction. Standard Flow Matching learns only the transport from a Gaussian source distribution to the distribution of artifact-free CT images and therefore cannot directly establish a correspondence between a random initial state and the current metal-affected sample. SCMA uses a linear-interpolation-corrected image to provide a sample-specific structural condition, employs dynamic spatial weighting over the metal region to enhance the learning of locally severe degradation, and constrains the generated result using the original projections during inference. In this way, a generic image-distribution prior is transformed into a conditional restoration model tailored to MAR.

\subsection{Preprocessing}
\label{subsec:preprocessing}

As shown in Fig.~\ref{fig:method_overview}(A), the preprocessing stage generates an image-domain metal mask, a projection-domain metal trace, and a linear-interpolation-corrected image. Following~\cite{kalender1987reduction}, we use the linear-interpolation-corrected image as the conditioning image and denote it by $\boldsymbol{x}_{\mathrm{Linear}}$. Given the original metal-affected projection $\boldsymbol{y}_{\mathrm{MA}}$ and its reconstructed image $\boldsymbol{x}_{\mathrm{MA}}$, we first employ a pretrained segmentation network $\mathcal{S}_{\phi}(\cdot)$ to obtain the image-domain metal mask:
\begin{equation}
\boldsymbol{M}
=
\mathcal{S}_{\phi}
\left(
\boldsymbol{x}_{\mathrm{MA}}
\right),
\qquad
M(\boldsymbol{p})\in\{0,1\},
\quad
\boldsymbol{p}\in\Omega,
\label{eq:method_metal_segmentation}
\end{equation}
where $\Omega$ denotes the image domain. Specifically, $M(\boldsymbol{p})=1$ indicates that pixel $\boldsymbol{p}$ belongs to a metal region, whereas $M(\boldsymbol{p})=0$ otherwise. This work does not develop a new metal segmentation model; instead, the mask $\boldsymbol{M}$ obtained using an existing segmentation model serves as a regional prior for subsequent processing.

The mask $\boldsymbol{M}$ is then forward-projected into the projection domain to obtain the binary metal trace:
\begin{equation}
\boldsymbol{T}
=
\mathbb{I}
\left[
\mathcal{F}
\left(
\boldsymbol{M}
\right)
>
0
\right],
\label{eq:method_metal_trace}
\end{equation}
where $\mathcal{F}(\cdot)$ denotes the CT forward-projection operator and $\mathbb{I}[\cdot]$ is the indicator function. Here, $T(\boldsymbol{r})=1$ indicates that the ray corresponding to projection-domain position $\boldsymbol{r}$ intersects a metal region.

Based on $\boldsymbol{T}$, the projection measurements within the metal trace are linearly interpolated using the nonmetal measurements on both sides of the trace. The resulting projection is then reconstructed using the reconstruction operator $\mathcal{R}(\cdot)$ to obtain the conditioning image:
\begin{equation}
\boldsymbol{x}_{\mathrm{Linear}}
=
\mathcal{R}
\left(
\left(
\boldsymbol{1}
-
\boldsymbol{T}
\right)
\odot
\boldsymbol{y}_{\mathrm{MA}}
+
\boldsymbol{T}
\odot
\mathcal{I}_{\mathrm{Linear}}
\left(
\boldsymbol{y}_{\mathrm{MA}},
\boldsymbol{T}
\right)
\right),
\label{eq:method_condition_image}
\end{equation}
where $\odot$ denotes element-wise multiplication and $\mathcal{I}_{\mathrm{Linear}}(\cdot)$ represents linear interpolation within the metal trace. The image $\boldsymbol{x}_{\mathrm{Linear}}$ may still contain interpolation errors, residual artifacts, and structural blurring. It is therefore not treated as the final corrected result but as a conditioning image that preserves the coarse structure of the current sample. The mask $\boldsymbol{M}$ is used for spatial weighting during training, whereas the corresponding metal trace $\boldsymbol{T}$ is used both to construct the conditioning image and to perform projection-consistency correction during inference. Consequently, metal segmentation errors may still affect the restoration performance of SCMA.

\subsection{Structure-Conditioned and Metal-Aware Flow Matching Training}
\label{subsec:fm_training}

As illustrated in Fig.~\ref{fig:method_overview}(B), the training stage learns a sample-specific conditional velocity field guided by the conditioning image and employs dynamic spatial weighting over the metal region to adjust the contributions of different pixels to the training objective. The artifact-free CT image, conditioning image, and metal mask in the training data are denoted by $\boldsymbol{x}_{1}$, $\boldsymbol{x}_{\mathrm{Linear}}$, and $\boldsymbol{M}$, respectively.

\subsubsection{Structure-Conditioned Flow Matching}
\label{subsubsec:structure_conditioned_fm}

Given an artifact-free target image $\boldsymbol{x}_{1}$, Gaussian noise $\boldsymbol{z}\sim\mathcal{N}(\boldsymbol{0},\boldsymbol{I})$, and a randomly sampled time $t\sim\mathcal{U}(0,1)$, where $\boldsymbol{I}$ denotes the identity matrix, we construct the intermediate state $\boldsymbol{x}_{t}$ according to Eq.~\eqref{eq:related_fm_path}, with its target velocity $\boldsymbol{u}_{t}$ given by Eq.~\eqref{eq:related_fm_velocity}. Under this definition, $t=0$ corresponds to the Gaussian-noise endpoint, whereas $t=1$ corresponds to the artifact-free image endpoint.

A standard Flow Matching velocity network $\boldsymbol{v}_{\theta}(\boldsymbol{x}_{t},t)$ predicts the transport direction using only the current state and time and therefore cannot determine the specific sample structure to be restored. To introduce a sample-specific condition, we concatenate $\boldsymbol{x}_{t}$ with its corresponding conditioning image $\boldsymbol{x}_{\mathrm{Linear}}$ along the channel dimension:
\begin{equation}
\boldsymbol{h}_{t}
=
\operatorname{Concat}
\left(
\boldsymbol{x}_{t},
\boldsymbol{x}_{\mathrm{Linear}}
\right),
\label{eq:method_condition_concat}
\end{equation}
and use the time-conditioned velocity network to predict
\begin{equation}
\widehat{\boldsymbol{u}}_{t}
=
\boldsymbol{v}_{\theta}
\left(
\boldsymbol{h}_{t},
t
\right).
\label{eq:method_conditional_velocity}
\end{equation}
Because $\boldsymbol{x}_{\mathrm{Linear}}$ remains fixed across all time states, it provides the coarse anatomical structure associated with the current metal-affected sample. The velocity network consequently learns a conditional restoration trajectory corresponding to the structure of the current sample, rather than an unconditional CT image generation process.

\subsubsection{Metal Mask-Guided Dynamic Weighting Loss}
\label{subsubsec:dynamic_weighted_loss}

Metal artifacts exhibit pronounced spatial nonuniformity. To increase the contribution of the metal region and its neighborhood to velocity-field training, we first compute a distance-transform map from the image-domain metal mask $\boldsymbol{M}$:
\begin{equation}
D(\boldsymbol{p})
=
\min_{\boldsymbol{q}:M(\boldsymbol{q})=1}
\left\|
\boldsymbol{p}
-
\boldsymbol{q}
\right\|_{2},
\qquad
\boldsymbol{p}\in\Omega,
\label{eq:method_distance_map}
\end{equation}
where $D(\boldsymbol{p})$ denotes the Euclidean distance from pixel $\boldsymbol{p}$ to the nearest metal region.

Based on $\boldsymbol{M}$ and $\boldsymbol{D}$, we construct a two-dimensional pixel-wise weight map $\boldsymbol{W}_{t}$ for each time point $t$:
\begin{equation}
W_{t}(\boldsymbol{p})
=
1
+
\alpha(t)M(\boldsymbol{p})
+
\beta(t)
\left[
1-M(\boldsymbol{p})
\right]
\exp
\left(
-\frac{D^{2}(\boldsymbol{p})}{2\sigma_{d}^{2}(t)}
\right),
\label{eq:method_weight_map}
\end{equation}
where $\alpha(t)\geq0$ and $\beta(t)\geq0$ control the weighting strengths within the metal region and its neighborhood, respectively, while $\sigma_{d}(t)$ controls the spatial extent of the neighborhood weighting. The neighborhood scale is defined as
\begin{equation}
\sigma_{d}(t)
=
(1-t)\sigma_{\max}
+
t\sigma_{\min},
\qquad
\sigma_{\max}>\sigma_{\min}>0.
\label{eq:method_time_dependent_sigma}
\end{equation}
Near the noise endpoint, $\boldsymbol{W}_{t}$ therefore assigns elevated weights to a broader neighborhood around the metal. As $t$ increases, the spatial extent of the elevated weights gradually contracts, allowing the model to focus more strongly on the metal region and its immediate neighborhood as the state approaches the artifact-free image endpoint. The weight maps shown at different times in Fig.~\ref{fig:method_overview}(B) represent separate two-dimensional weight maps rather than multiple feature channels at the same time point.

Using the predicted velocity obtained from Eq.~\eqref{eq:method_conditional_velocity}, we define the velocity prediction error at time $t$ and pixel position $\boldsymbol{p}$ as
\begin{equation}
\boldsymbol{E}_{t}(\boldsymbol{p})
=
\widehat{\boldsymbol{u}}_{t}(\boldsymbol{p})
-
\boldsymbol{u}_{t}(\boldsymbol{p}),
\qquad
\boldsymbol{p}\in\Omega.
\label{eq:pixel_error}
\end{equation}
The dynamically spatially weighted Flow Matching loss of SCMA is defined as
\begin{equation}
\mathcal{L}_{\mathrm{wFM}}
=
\mathbb{E}_{\substack{
(\boldsymbol{x}_{1},\boldsymbol{x}_{\mathrm{Linear}},\boldsymbol{M})
\sim\mathcal{P}_{\mathrm{train}},\\
\boldsymbol{z}\sim\mathcal{N}(\boldsymbol{0},\boldsymbol{I}),
\;t\sim\mathcal{U}(0,1)
}}
\left[
\frac{
\displaystyle
\sum_{\boldsymbol{p}\in\Omega}
W_{t}(\boldsymbol{p})
\left\|
\boldsymbol{E}_{t}(\boldsymbol{p})
\right\|_{2}^{2}
}{
\displaystyle
\sum_{\boldsymbol{p}\in\Omega}
W_{t}(\boldsymbol{p})
+
\varepsilon
}
\right],
\label{eq:method_weighted_fm_loss}
\end{equation}
where $\mathcal{P}_{\mathrm{train}}$ denotes the training-sample distribution and $\varepsilon$ is a numerical stability term. The denominator reduces the effect of variations in the overall scale of different weight maps on the loss magnitude. Compared with a spatially uniform Flow Matching loss, Eq.~\eqref{eq:method_weighted_fm_loss} increases the relative contribution of velocity prediction errors within and around the metal region, enabling the network to learn more effectively from locally severe degradation.

Algorithm~\ref{alg:SCMA_training} summarizes the preprocessing and training procedures of SCMA.

\begin{algorithm}[t]
\caption{Preprocessing and Training of SCMA}
\label{alg:SCMA_training}
\small
\begin{algorithmic}[1]
\STATE \textbf{Input:} metal-affected image $\boldsymbol{x}_{\mathrm{MA}}$, projection $\boldsymbol{y}_{\mathrm{MA}}$, and artifact-free target $\boldsymbol{x}_{1}$
\STATE Obtain the metal mask $\boldsymbol{M}$ using Eq.~\eqref{eq:method_metal_segmentation}
\STATE Generate the metal trace $\boldsymbol{T}$ using Eq.~\eqref{eq:method_metal_trace}
\STATE Compute the conditioning image $\boldsymbol{x}_{\mathrm{Linear}}$ using Eq.~\eqref{eq:method_condition_image}
\FOR{each training iteration}
    \STATE Sample $(\boldsymbol{x}_{1},\boldsymbol{x}_{\mathrm{Linear}},\boldsymbol{M})\sim\mathcal{P}_{\mathrm{train}}$
    \STATE Sample $\boldsymbol{z}\sim\mathcal{N}(\boldsymbol{0},\boldsymbol{I})$ and $t\sim\mathcal{U}(0,1)$
    \STATE Compute $\boldsymbol{x}_{t}$ and $\boldsymbol{u}_{t}$ using Eqs.~\eqref{eq:related_fm_path} and~\eqref{eq:related_fm_velocity}
    \STATE Construct $\boldsymbol{h}_{t}$ using Eq.~\eqref{eq:method_condition_concat}
    \STATE Compute $\boldsymbol{D}$ and $\boldsymbol{W}_{t}$ using Eqs.~\eqref{eq:method_distance_map}--\eqref{eq:method_time_dependent_sigma}
    \STATE Predict $\widehat{\boldsymbol{u}}_{t}=\boldsymbol{v}_{\theta}(\boldsymbol{h}_{t},t)$
    \STATE Compute $\boldsymbol{E}_{t}(\boldsymbol{p})=\widehat{\boldsymbol{u}}_{t}(\boldsymbol{p})-\boldsymbol{u}_{t}(\boldsymbol{p})$
    \STATE Update $\theta$ by minimizing $\mathcal{L}_{\mathrm{wFM}}$ in Eq.~\eqref{eq:method_weighted_fm_loss}
\ENDFOR
\STATE \textbf{return} the trained structure-conditioned velocity network $\boldsymbol{v}_{\theta}$
\end{algorithmic}
\end{algorithm}

\subsection{Projection-Consistency Correction Inference}
\label{subsec:pcc_inference}

As shown in Fig.~\ref{fig:method_overview}(C), the inference stage solves the learned Flow Matching ordinary differential equation under the guidance of the fixed conditioning image $\boldsymbol{x}_{\mathrm{Linear}}$ and performs projection-consistency correction (PCC) at each time step using the original projection measurements outside the metal trace. Let the discrete inference time sequence be
\begin{equation}
0
=
\tau_{0}
<
\tau_{1}
<
\cdots
<
\tau_{K}
=
1,
\label{eq:method_time_sequence}
\end{equation}
where $K$ denotes the number of inference steps. The initial state is
\begin{equation}
\boldsymbol{x}_{\tau_{0}}
=
\boldsymbol{z},
\qquad
\boldsymbol{z}
\sim
\mathcal{N}
\left(
\boldsymbol{0},
\boldsymbol{I}
\right).
\label{eq:method_initial_noise}
\end{equation}

\begin{algorithm}[t]
\caption{SCMA Inference with Projection-Consistency Correction}
\label{alg:SCMA_inference}
\small
\begin{algorithmic}[1]
\STATE \textbf{Input:} trained network $\boldsymbol{v}_{\theta}$, conditioning image $\boldsymbol{x}_{\mathrm{Linear}}$, projection $\boldsymbol{y}_{\mathrm{MA}}$, metal trace $\boldsymbol{T}$, and time sequence $\{\tau_k\}_{k=0}^{K}$
\STATE Sample $\boldsymbol{x}_{\tau_{0}}=\boldsymbol{z}$, where $\boldsymbol{z}\sim\mathcal{N}(\boldsymbol{0},\boldsymbol{I})$
\FOR{$k=0,\ldots,K-1$}
    \STATE Construct $\boldsymbol{h}_{\tau_k}=\operatorname{Concat}(\boldsymbol{x}_{\tau_k},\boldsymbol{x}_{\mathrm{Linear}})$
    \STATE Predict $\widehat{\boldsymbol{u}}_{\tau_k}=\boldsymbol{v}_{\theta}(\boldsymbol{h}_{\tau_k},\tau_k)$
    \STATE Estimate $\widehat{\boldsymbol{x}}_{1,\tau_k}$ using Eq.~\eqref{eq:method_x0_prediction}
    \STATE Compute $\boldsymbol{x}_{\mathrm{PCC},\tau_k}$ using Eq.~\eqref{eq:method_projection_dc}
    \STATE Update $\boldsymbol{x}_{\tau_{k+1}}$ using Eq.~\eqref{eq:method_state_update}
\ENDFOR
\STATE \textbf{return} $\boldsymbol{x}_{\tau_K}$
\end{algorithmic}
\end{algorithm}

At the $k$th inference stage, $\boldsymbol{x}_{\tau_k}$ is the current state updated throughout the inference process, whereas the conditioning image $\boldsymbol{x}_{\mathrm{Linear}}$ remains fixed across all time steps. They are concatenated along the channel dimension:
\begin{equation}
\boldsymbol{h}_{\tau_{k}}
=
\operatorname{Concat}
\left(
\boldsymbol{x}_{\tau_{k}},
\boldsymbol{x}_{\mathrm{Linear}}
\right),
\label{eq:method_inference_condition}
\end{equation}
and the trained conditional velocity network predicts
\begin{equation}
\widehat{\boldsymbol{u}}_{\tau_{k}}
=
\boldsymbol{v}_{\theta}
\left(
\boldsymbol{h}_{\tau_{k}},
\tau_{k}
\right).
\label{eq:method_sampling_velocity}
\end{equation}
According to the linear probability path, the artifact-free endpoint corresponding to the current state is estimated as
\begin{equation}
\widehat{\boldsymbol{x}}_{1,\tau_{k}}
=
\boldsymbol{x}_{\tau_{k}}
+
\left(
1-\tau_{k}
\right)
\widehat{\boldsymbol{u}}_{\tau_{k}}.
\label{eq:method_x0_prediction}
\end{equation}

Because the original measurements within the metal trace are severely corrupted, only the relatively reliable projections outside the metal trace are used to constrain the endpoint estimate. The reliable-region mask in the projection domain is defined as
\begin{equation}
\boldsymbol{W}_{\mathrm{p}}
=
\boldsymbol{1}
-
\boldsymbol{T}.
\label{eq:method_projection_weight}
\end{equation}
The PCC problem at the $k$th inference stage is formulated as
\begin{equation}
\begin{aligned}
\boldsymbol{x}_{\mathrm{PCC},\tau_{k}}
=
\arg\min_{\boldsymbol{x}}
\quad
&
\frac{1}{2}
\left\|
\boldsymbol{W}_{\mathrm{p}}
\odot
\left[
\mathcal{F}(\boldsymbol{x})
-
\boldsymbol{y}_{\mathrm{MA}}
\right]
\right\|_{2}^{2}
\\
&
+
\frac{\rho_{k}}{2}
\left\|
\boldsymbol{x}
-
\widehat{\boldsymbol{x}}_{1,\tau_{k}}
\right\|_{2}^{2},
\end{aligned}
\label{eq:method_projection_dc}
\end{equation}
where $\rho_{k}>0$ controls the trade-off between projection consistency and the Flow Matching image prior. The first term enforces consistency between the corrected result and the original measurements outside the metal trace, while the second prevents the corrected result from deviating excessively from the current endpoint estimate. For notational simplicity, the solution of Eq.~\eqref{eq:method_projection_dc} is expressed as
\begin{equation}
\boldsymbol{x}_{\mathrm{PCC},\tau_{k}}
=
\mathcal{C}_{\mathrm{PCC}}
\left(
\widehat{\boldsymbol{x}}_{1,\tau_{k}};
\boldsymbol{y}_{\mathrm{MA}},
\boldsymbol{T},
\rho_{k}
\right).
\label{eq:method_dc_operator}
\end{equation}

The correction produced by PCC for the endpoint estimate is subsequently fed back into the current Flow Matching state, and the state at the next time point is obtained using a first-order Euler update:
\begin{equation}
\begin{aligned}
\boldsymbol{x}_{\tau_{k+1}}
=
\boldsymbol{x}_{\tau_{k}}
&+
\left(
\tau_{k+1}
-
\tau_{k}
\right)
\widehat{\boldsymbol{u}}_{\tau_{k}}
\\
&+
\omega_{k}
\left(
\boldsymbol{x}_{\mathrm{PCC},\tau_{k}}
-
\widehat{\boldsymbol{x}}_{1,\tau_{k}}
\right),
\end{aligned}
\label{eq:method_state_update}
\end{equation}
where $0\leq\omega_{k}\leq1$ denotes the correction feedback strength. The updated state $\boldsymbol{x}_{\tau_{k+1}}$ is fed back into the velocity network as the variable state for the next inference stage, while $\boldsymbol{x}_{\mathrm{Linear}}$ remains the fixed condition. This process jointly constrains the inference trajectory using the sample-specific structural condition, the artifact-free image-distribution prior, and reliable projection measurements. After the iteration reaches $\tau_K=1$, $\boldsymbol{x}_{\tau_K}$ is taken as the final MAR result.

Algorithm~\ref{alg:SCMA_inference} summarizes the SCMA inference procedure with PCC.

\section{Experiments}

\begin{table*}[t]
  \centering
  \caption{Quantitative comparison of different MAR methods for small, medium, and large metal implants. The average PSNR (dB) $\uparrow$ and SSIM (\%) $\uparrow$ are reported. The best and second-best results are highlighted in bold and underlined, respectively.}
  \label{tab:quantitative_comparison}
  \footnotesize
  \setlength{\tabcolsep}{5.2pt}
  \renewcommand{\arraystretch}{1.15}
  \begin{tabular}{lcccccccc}
    \toprule
    \multirow{2}{*}{Method} & \multicolumn{2}{c}{Small metal} & \multicolumn{2}{c}{Medium metal} & \multicolumn{2}{c}{Large metal} & \multicolumn{2}{c}{Average} \\
    \cmidrule(lr){2-3} \cmidrule(lr){4-5} \cmidrule(lr){6-7} \cmidrule(lr){8-9}
    & PSNR & SSIM & PSNR & SSIM & PSNR & SSIM & PSNR & SSIM \\
    \midrule
    FBP        & 18.1664 & 60.3932 & 17.6839 & 77.4033 & 5.1595  & 33.9951 & 13.6699 & 57.2639 \\
    NMAR       & 39.6144 & 86.7410 & 39.1023 & 93.3539 & 37.3240 & 85.0119 & 38.6802 & 88.3689 \\
    DICDNet    & \underline{50.0476} & \underline{99.5179} & \underline{43.5349} & \underline{99.2870} & \underline{41.6628} & \underline{97.7062} & \underline{45.0818} & \underline{98.8370} \\
    InDuDoNet+ & 47.3177 & 99.0293 & 42.3334 & 99.2426 & 34.7378 & 96.1927 & 41.4630 & 98.1549 \\
    CALIMAR    & 35.4715 & 87.9920 & 38.6385 & 93.3364 & 32.5506 & 83.9158 & 35.5535 & 88.4148 \\
    ADN        & 35.8166 & 95.4741 & 36.9620 & 96.8683 & 27.8158 & 88.0662 & 33.5315 & 93.4696 \\
    DuDoDp-MAR & 36.1534 & 85.6365 & 39.3705 & 92.9689 & 34.7557 & 84.0931 & 36.7599 & 87.5662 \\
    Proposed   & \textbf{51.0738} & \textbf{99.6571} & \textbf{48.8946} & \textbf{99.4946} & \textbf{42.7066} & \textbf{97.9768} & \textbf{47.5583} & \textbf{99.0429} \\
    \bottomrule
  \end{tabular}
\end{table*}

\subsection{Experimental Setup}
\label{subsec:experimental_setup}

\begin{figure*}[!t]
    \centering
    \includegraphics[width=\textwidth]{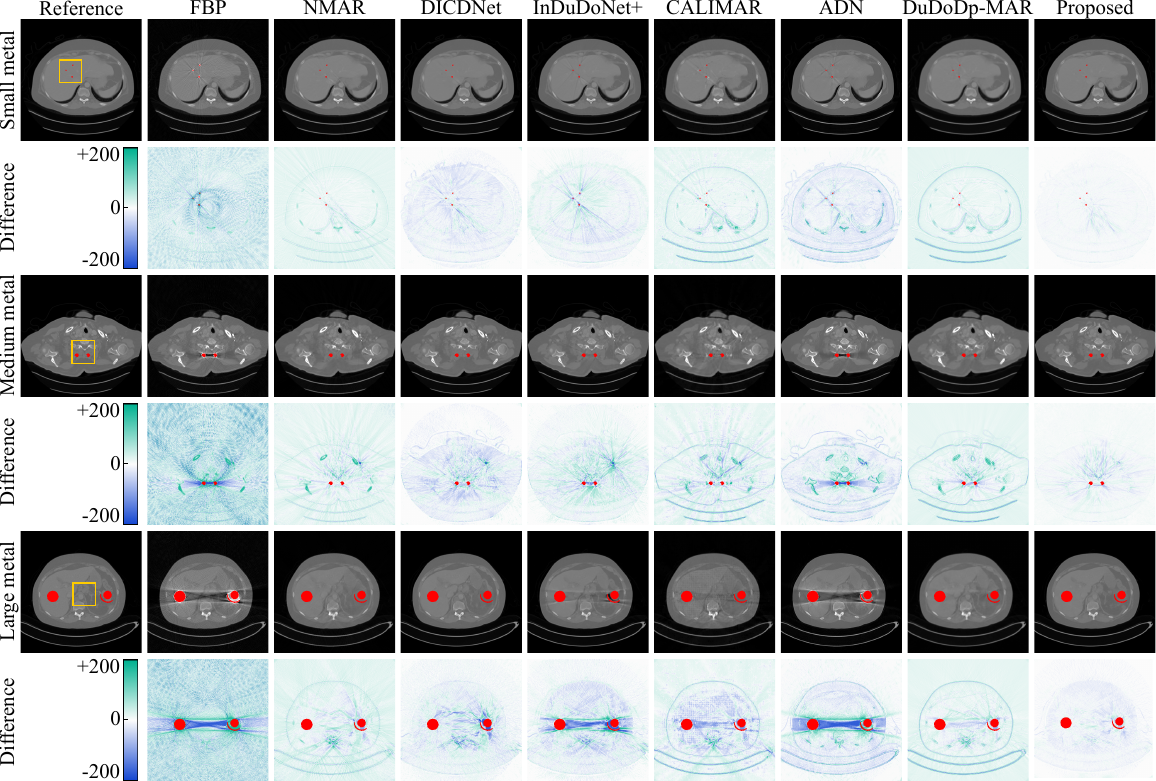}
    \caption{Visual comparison of different MAR methods for small, medium, and large metal implants. The columns from left to right show Reference, FBP, NMAR, DICDNet, InDuDoNet+, CALIMAR, ADN, DuDoDp-MAR, and the proposed SCMA. For each metal-size category, the upper row shows the corrected CT images displayed within $[\mathrm{-1000},\mathrm{800}]$ HU, and the lower row shows the residual maps relative to the reference images displayed within $[-200,200]$ HU. The metal regions are marked in red.}
    \label{fig:comparison}
\end{figure*}

\subsubsection{Datasets}
\label{subsubsec:datasets}

Following the data simulation protocols adopted by existing MAR methods~\cite{liao2020adn,wang2022dicdnet,wang2023indudonet}, we randomly selected 1,200 metal-free CT images from the DeepLesion dataset~\cite{yan2018deep} and collected 100 metal masks with different sizes and shapes to synthesize metal-affected images. Among them, 90 metal masks and 1,000 metal-free CT images were used to synthesize the training samples. The remaining 10 metal masks and 200 metal-free CT images were used to construct the test set, yielding 2,000 paired metal-affected and metal-free test images. The sizes of the 10 test metal implants were $[32, 53, 112, 115, 115, 242, 448, 878, 879, 2054]$ pixels. According to the metal size, the test samples were divided into small-, medium-, and large-metal categories to evaluate MAR performance under different metal sizes. Specifically, small metal implants were defined as those containing $[0,100]$ pixels, medium metal implants as those containing $[101,500]$ pixels, and large metal implants as those containing more than 500 pixels.

All CT images were resized to $416\times416$ pixels with a pixel spacing of $0.369\,\mathrm{mm}$. Projection data were simulated using a two-dimensional fan-beam CT geometry. The source-to-object distance (SOD) and source-to-detector distance (SDD) were $396.92\,\mathrm{mm}$ and $793.85\,\mathrm{mm}$, respectively. The one-dimensional detector contained 641 detector elements with a total width of $434.45\,\mathrm{mm}$, corresponding to a detector-element width of approximately $0.678\,\mathrm{mm}$. For each sample, 640 projection views were uniformly distributed over the range of $0$--$360^{\circ}$. Following the same simulation procedure, we generated a metal-affected image, an LI-corrected image, a metal mask, a metal trace, and the corresponding metal-free reference image for each sample.

To further evaluate the generalization capability of the proposed method in real metal-artifact scenarios, we selected three clinical CT volumes containing real metal implants from the COLONOG subset of the CTSpine1K dataset~\cite{deng2025ctspine1k}. The case identifiers were 0064, 0131, and 0313. This subset was originally derived from the publicly available CT COLONOGRAPHY data in The Cancer Imaging Archive (TCIA). The original in-plane image size of all three cases was $512\times512$ pixels, with 507, 663, and 554 slices, respectively. Their interslice spacing was $0.8\,\mathrm{mm}$, while their in-plane pixel spacing was $0.703\,\mathrm{mm}$, $0.781\,\mathrm{mm}$, and $0.859\,\mathrm{mm}$, respectively. We extracted 72, 153, and 175 axial slices containing metal implants and associated artifacts from these three volumes, yielding a total of 400 real metal-affected CT images. It should be emphasized that these images were obtained from three three-dimensional cases rather than 400 independent patients.

To match the network input size, all slices were resized to $416\times416$ pixels while preserving the complete field of view. A pretrained nnU-Net was employed to generate the corresponding image-domain metal masks. Across all 400 slices, the segmented metal-region sizes ranged from 29 to 2,099 pixels. Using the same metal-size criteria as those applied to the synthetic test set, the 400 valid metal-containing slices comprised 10 small-metal samples, 143 medium-metal samples, and 247 large-metal samples.

\subsubsection{Training Details}
\label{subsubsec:training_details}

The proposed SCMA method was implemented in PyTorch. The conditional Flow Matching framework was trained on an NVIDIA GeForce RTX 5080 GPU. Its input was constructed by concatenating the current intermediate state $\boldsymbol{x}_{t}$ and the linear-interpolation-corrected image $\boldsymbol{x}_{\mathrm{Linear}}$ along the channel dimension, i.e., $(\boldsymbol{x}_{t},\boldsymbol{x}_{\mathrm{Linear}})$ was used as the conditional input.

The network was optimized using AdamW with the default momentum parameters in PyTorch. The initial learning rate was set to $5\times10^{-5}$, the weight decay was set to 0, and the gradient-clipping threshold was set to 1.0. The batch size was set to 1, and the number of Flow Matching training time steps was set to 1,000. A conditional U-Net with 64 base channels was employed as the backbone, with FFT Transformer modules incorporated into its low-resolution feature levels. At each training iteration, one metal-free CT image and one synthetic metal mask were randomly selected from the pool of 1,000 training images and the pool of 90 training masks, respectively, to synthesize a metal-affected CT image.

\subsection{Comparison With State-of-the-Art Methods}
\label{subsec:comparison_existing_methods}

\begin{figure*}[!t]
    \centering
    \includegraphics[width=\textwidth]{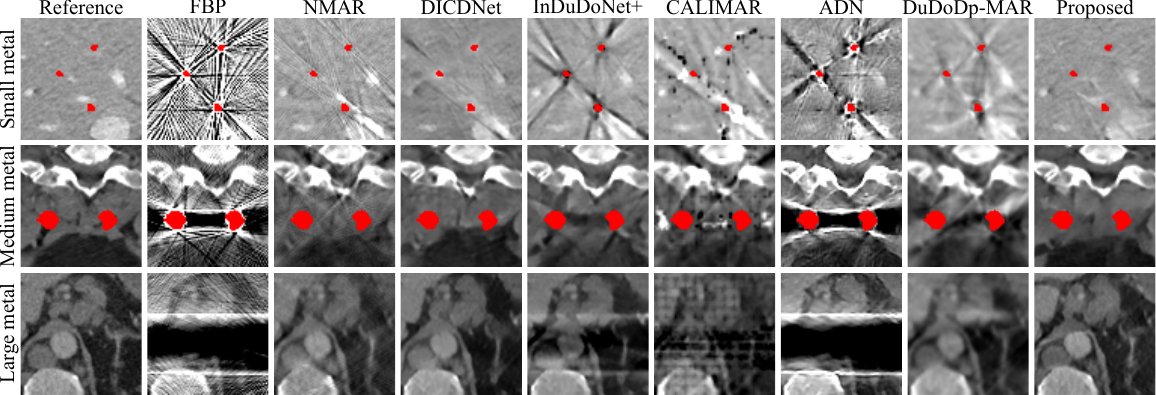}
    \caption{Local ROI comparison of different MAR methods for small, medium, and large metal implants. The ROIs in the first row are displayed within $[\mathrm{-250},\mathrm{50}]$ HU, whereas those in the second and third rows are displayed within $[\mathrm{-350},\mathrm{100}]$ HU. The metal regions are marked in red. The proposed SCMA more effectively suppresses residual streak and dark-band artifacts around the metal regions while preserving the soft-tissue background and bone boundaries.}
    \label{fig:roi_comparison}
\end{figure*}

We compared SCMA with representative methods from several major categories of MAR techniques. In addition to FBP as the analytical reconstruction baseline, we selected NMAR~\cite{meyer2010normalized} as a model-based projection-domain completion method; DICDNet~\cite{wang2022dicdnet} and InDuDoNet+~\cite{wang2023indudonet} as supervised deep MAR methods; CALIMAR~\cite{scardigno2025calimargan} and ADN~\cite{liao2020adn} as unpaired or unsupervised MAR methods; and DuDoDp-MAR~\cite{liu2024diffusionpriors} as a generative-prior-based MAR method. These methods cover the principal technical paradigms and representative advanced approaches for CT metal artifact reduction. All methods were evaluated on the same test set using PSNR and SSIM as the quantitative metrics.

Table~\ref{tab:quantitative_comparison} presents the quantitative results for different metal sizes. Because FBP does not correct the projection inconsistencies caused by metal, its overall performance is substantially limited. NMAR considerably reduces streak artifacts but remains affected by interpolation errors and the quality of its prior image. The deep learning-based methods generally outperform the conventional methods, with DICDNet achieving the most competitive performance among the compared methods. In contrast, SCMA achieves the best performance for small, medium, and large metal implants, with average PSNR and SSIM values of $47.5583$ dB and $99.0429\%$, respectively. Compared with the second-best method, DICDNet, SCMA improves the average PSNR by $2.4765$ dB, with a particularly pronounced advantage for medium-sized metal implants. These results demonstrate that LI-based conditioning, dynamically weighted training over the metal region, and projection-consistency correction effectively improve restoration robustness across different levels of artifact severity.

Figure~\ref{fig:comparison} presents the visual comparison of the different methods. FBP exhibits pronounced radial streaks and dark-band artifacts, whose severity increases with the metal size. Although NMAR removes some severe artifacts, residual errors and interpolation-induced secondary artifacts remain visible around the metal regions. DICDNet and InDuDoNet+ substantially improve the overall image quality but retain structured residuals near the metal regions and high-contrast structural boundaries. CALIMAR, ADN, and DuDoDp-MAR further reduce some artifacts, but local intensity shifts, abnormal textures, or structural oversmoothing can still be observed. In comparison, SCMA produces weaker and more localized residual responses and exhibits greater stability around the metal regions, in soft-tissue backgrounds, and along bone boundaries. The local ROI comparison in Fig.~\ref{fig:roi_comparison} further shows that SCMA preserves the continuity and clarity of local anatomical structures while suppressing severe artifacts, consistent with the quantitative results.

\subsection{Experiments on Real-World Data}

\begin{figure*}[!t]
    \centering
    \includegraphics[width=0.897\textwidth]{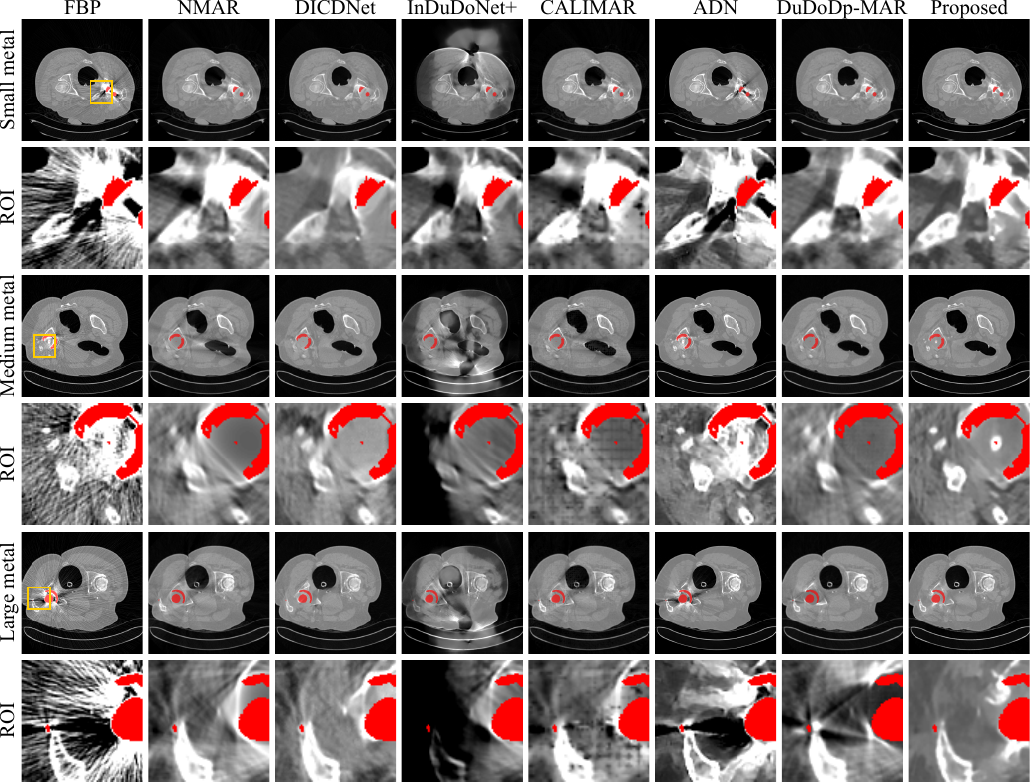}
    \caption{Visual comparison of different MAR methods on real-world CT data containing small, medium, and large metal implants. The columns from left to right show FBP, NMAR, DICDNet, InDuDoNet+, CALIMAR, ADN, DuDoDp-MAR, and the proposed SCMA. For each metal-size category, the upper row shows the corrected CT images displayed within $[\mathrm{-1000},\mathrm{800}]$ HU, and the lower row shows the corresponding metal-adjacent ROIs displayed within $[-350,250]$ HU. The metal regions are marked in red.}
    \label{fig:experiment}
\end{figure*}

To further evaluate the generalization capability of SCMA under real acquisition conditions, we conducted experiments on real-world CT data containing small, medium, and large metal implants and compared the results with those obtained using FBP, NMAR, DICDNet, InDuDoNet+, CALIMAR, ADN, and DuDoDp-MAR. Because strictly registered metal-free reference images were unavailable for the real-world data, the different methods were evaluated qualitatively using both the complete CT images and the metal-adjacent ROIs. The same display windows were used for all methods to ensure a fair visual comparison.

As shown in Fig.~\ref{fig:experiment}, the radial streaks and bright--dark distortions in the FBP images become substantially more severe as the metal size increases, strongly interfering with tissue structures around the metal regions. NMAR, DICDNet, CALIMAR, and DuDoDp-MAR reduce the global streak artifacts to some extent, but residual streaks, shading artifacts, or structural blurring remain visible in the magnified regions. InDuDoNet+ produces noticeable intensity nonuniformity and structural distortion in some regions, particularly in the medium- and large-metal cases. Although ADN restores some tissue information, pronounced intensity variations and secondary artifacts remain around the metal regions.

In comparison, SCMA provides more stable artifact correction across all three metal sizes. For the small metal implant, SCMA effectively suppresses the surrounding radial streaks while preserving the boundaries of adjacent tissues. For the medium-sized implant, it produces a more uniform intensity distribution and clearer local tissue structures. In the more challenging large-metal case, SCMA still substantially reduces bright--dark streaks and shading artifacts around the metal region while preserving the continuity of bone structures and soft-tissue boundaries. These results demonstrate that SCMA adapts well to different metal sizes and achieves a favorable balance between artifact suppression and local structure preservation on real-world data.

\subsection{Generalization Across Metal Sizes}
\label{subsec:generalization_metal_size}

\begin{figure}[!t]
    \centering
    \includegraphics[width=\columnwidth]{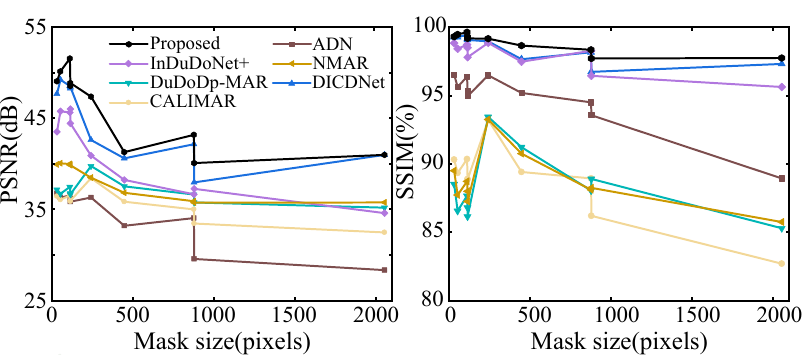}
    \caption{Comparison of the generalization performance of different MAR methods across metal sizes. The horizontal axis indicates the number of pixels in each metal mask, while the vertical axes indicate PSNR (dB) $\uparrow$ and SSIM (\%) $\uparrow$, respectively. For each metal mask, 10 test slices were randomly selected to synthesize metal-affected images, and the average PSNR and SSIM between the corrected and reference images were calculated for each method.}
    \label{fig:generalization_metal_size}
\end{figure}

To further evaluate the generalization capability of different methods across varying metal sizes, we selected 10 metal masks containing 32, 53, 112, 115, 115, 242, 448, 878, 879, and 2,054 pixels, respectively. For each metal mask, 10 test slices were randomly selected to synthesize metal-affected images. The average PSNR and SSIM between the corrected and reference images were then calculated for each method. The results are presented in Fig.~\ref{fig:generalization_metal_size}.

In general, as the metal size increases, more projection information within the metal traces becomes missing or unreliable, thereby increasing the difficulty of MAR. Consequently, both PSNR and SSIM exhibit an overall decreasing trend. In comparison, SCMA consistently maintains high reconstruction quality across different metal sizes and generally outperforms the competing methods in terms of both PSNR and SSIM, demonstrating improved restoration robustness across different metal scales. DICDNet and InDuDoNet+ also perform well for some metal sizes but exhibit performance degradation in cases involving larger or more complex metal implants. The curves of CALIMAR, ADN, DuDoDp-MAR, and NMAR fluctuate more substantially, indicating greater sensitivity to metal size, location, and local anatomical complexity.

It should be noted that MAR difficulty is not determined monotonically by metal-mask area alone but is also affected by metal shape, location, and the complexity of the occluded anatomy. For example, the average PSNR and SSIM obtained for the 878-pixel metal mask are higher than those obtained for the 448-pixel mask. Although the former has a larger area, it is primarily located within a relatively homogeneous tissue region, making it easier to recover structures close to the reference after metal removal. In contrast, the smaller 448-pixel mask occludes a more complex anatomical region and therefore results in greater restoration difficulty and lower average metric values. Similarly, although the 878- and 879-pixel metal masks have nearly identical areas, their performance differs substantially because the 878-pixel metal has a relatively simple structure, whereas the 879-pixel metal has a more complex shape, leading to markedly different artifact distributions and restoration difficulties. A similar phenomenon is observed for the two 115-pixel metal masks, indicating that identical or similar metal areas do not necessarily correspond to the same level of MAR difficulty. These results demonstrate that SCMA generalizes not only across metal sizes but also across variations in metal shape and local anatomy.

\subsection{Local Structural Fidelity Analysis}
\label{subsec:local_structural_fidelity}

\begin{figure}[!t]
    \centering
    \includegraphics[width=\columnwidth]{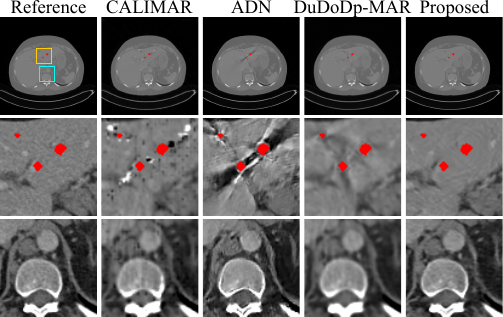}
    \caption{Comparison of hallucination-like structures produced by generative and unpaired MAR methods. The columns from left to right show Reference, CALIMAR, ADN, DuDoDp-MAR, and the proposed SCMA. The first row shows the complete CT images displayed within $[-1000,800]$ HU. The second and third rows show the metal-adjacent ROI indicated by the yellow box and the spine ROI indicated by the cyan box, respectively, both displayed within $[-350,100]$ HU. The metal regions are marked in red.}
    \label{fig:hallucination}
\end{figure}

\begin{table}[htbp]
  \centering
  \caption{Quantitative results for local structural fidelity within two ROIs obtained using different generative and unpaired MAR methods. SSIM $\uparrow$ measures structural similarity, whereas HFEN $\downarrow$ measures high-frequency structural errors. The best results are highlighted in bold.}
  \label{tab:hallucination_roi_metrics}
  \footnotesize
  \setlength{\tabcolsep}{5.5pt}
  \renewcommand{\arraystretch}{1.15}
  \begin{tabular}{ccccc}
    \toprule
    \multirow{2}{*}{Method}
    & \multicolumn{2}{c}{Metal-adjacent ROI}
    & \multicolumn{2}{c}{Spine ROI} \\
    \cmidrule(lr){2-3} \cmidrule(lr){4-5}
    & SSIM & HFEN & SSIM & HFEN \\
    \midrule
    CALIMAR
    & 86.5036 & 0.9291
    & 91.1514 & 0.3583 \\
    ADN
    & 77.7021 & 1.1265
    & 91.9477 & 0.3344 \\
    DuDoDp-MAR
    & 93.6403 & 0.3939
    & 93.3840 & 0.2658 \\
    Proposed
    & \textbf{97.0522} & \textbf{0.2367}
    & \textbf{96.9391} & \textbf{0.1429} \\
    \bottomrule
  \end{tabular}
\end{table}

Generative priors can improve the ability of MAR models to represent the distribution of artifact-free CT images. However, in regions with severe information loss, such as those adjacent to metal implants, they may also generate hallucination-like structures that are inconsistent with the true anatomy. We therefore evaluated the local structural reliability of different generative or unpaired MAR methods using both visual results and quantitative ROI metrics. Specifically, SSIM was used to measure local structural similarity, whereas HFEN was used to quantify errors in high-frequency structures such as edges and textures~\cite{ravishankar2011mr}.

As shown in Fig.~\ref{fig:hallucination}, although CALIMAR and ADN reduce some global artifacts, abnormal bright--dark textures and streak-like residuals remain around the metal regions. DuDoDp-MAR improves the overall visual quality but still exhibits local oversmoothing and detail deviations. In comparison, SCMA more effectively suppresses spurious textures around the metal region in the metal-adjacent ROI while better preserving bone boundaries and the surrounding soft-tissue background in the spine ROI. The quantitative results in Table~\ref{tab:hallucination_roi_metrics} further support these observations. SCMA achieves the highest SSIM and lowest HFEN in both ROIs, indicating that its restored local structures are more consistent with the reference and contain smaller high-frequency structural errors.

\subsection{Ablation Studies}
\label{subsec:ablation_studies}

\begin{table*}[ht]
  \centering
  \caption{Quantitative ablation results for different components of SCMA under small-, medium-, and large-metal conditions. PSNR (dB) $\uparrow$ and SSIM (\%) $\uparrow$ are reported. The best results are highlighted in bold.}
  \label{tab:ablation}
  \footnotesize
  \setlength{\tabcolsep}{5.2pt}
  \renewcommand{\arraystretch}{1.15}
  \begin{tabular}{lccc cccccc cc}
    \toprule
    \multirow{2}{*}{Variant} & \multirow{2}{*}{LI} & \multirow{2}{*}{PCC} & \multirow{2}{*}{DWL} 
    & \multicolumn{2}{c}{Small metal} 
    & \multicolumn{2}{c}{Medium metal} 
    & \multicolumn{2}{c}{Large metal} 
    & \multicolumn{2}{c}{Average} \\
    \cmidrule(lr){5-6} \cmidrule(lr){7-8} \cmidrule(lr){9-10} \cmidrule(lr){11-12}
    & & & & PSNR & SSIM & PSNR & SSIM & PSNR & SSIM & PSNR & SSIM \\
    \midrule
    W/o LI  & $\times$ & $\checkmark$ & $\checkmark$ 
    & 17.4651 & 41.1641 
    & 16.6351 & 25.9686 
    & 16.3352 & 29.2517 
    & 16.8118 & 32.1281 \\
    W/o PCC & $\checkmark$ & $\times$ & $\checkmark$ 
    & 45.0270 & 89.8063 
    & 42.8753 & 86.3775 
    & 39.4970 & 85.6275 
    & 42.4664 & 87.2704 \\
    W/o DWL & $\checkmark$ & $\checkmark$ & $\times$ 
    & 47.2420 & 95.7863 
    & 43.5153 & 93.1681 
    & 37.6677 & 89.6246 
    & 42.8083 & 92.8597 \\
    Proposed & $\checkmark$ & $\checkmark$ & $\checkmark$ 
    & \textbf{51.0738} & \textbf{99.6571} 
    & \textbf{48.8946} & \textbf{99.4946} 
    & \textbf{42.7066} & \textbf{97.9768} 
    & \textbf{47.5583} & \textbf{99.0429} \\
    \bottomrule
  \end{tabular}
\end{table*}

\begin{figure}[!t]
    \centering
    \includegraphics[width=\columnwidth]{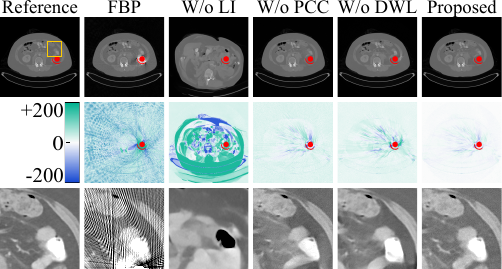}
    \caption{Visual ablation comparison of the different components of the proposed SCMA. The columns from left to right show Reference, FBP, W/o LI, W/o PCC, W/o DWL, and the complete SCMA. The first row shows the corrected CT images displayed within $[-1000,800]$ HU; the second row shows the residual maps relative to the reference image displayed within $[-200,200]$ HU; and the third row shows the local ROIs displayed within $[-450,100]$ HU. The metal regions are marked in red.}
    \label{fig:ablation}
\end{figure}

To evaluate the effectiveness of the individual components of SCMA, we constructed three ablation variants: W/o LI, which removes the linear-interpolation-based structural condition; W/o PCC, which removes projection-consistency correction; and W/o DWL, which removes the dynamic weighting loss. Except for the component being removed, all variants used identical training and testing settings. The quantitative and visual results are presented in Table~\ref{tab:ablation} and Fig.~\ref{fig:ablation}, respectively.

As shown in Table~\ref{tab:ablation}, removing any component results in performance degradation. W/o LI exhibits the most substantial deterioration, indicating that the coarse structural condition provided by the LI-corrected image is essential for stabilizing the Flow Matching inference trajectory. After PCC is removed, the model can still recover the principal structures using the image-domain generative prior, but its quantitative performance decreases considerably. This result indicates that relying solely on an image-domain prior cannot ensure consistency with the original projection measurements. Removing DWL weakens the model's restoration capability within and around the metal regions, demonstrating that a spatially uniform training objective is insufficient for adequately learning from regions affected by locally severe artifacts. In comparison, the complete SCMA achieves the best performance across all metal sizes, confirming the complementary roles of LI, PCC, and DWL.

Figure~\ref{fig:ablation} further illustrates the local differences among the variants. W/o LI exhibits pronounced global structural distortion and widespread residual errors, indicating that the generative process may deviate from the anatomy of the current sample in the absence of structural conditioning. W/o PCC and W/o DWL preserve the overall structure more effectively but retain more pronounced residual artifacts and local intensity deviations around the metal regions. The complete SCMA produces the weakest residual response, and its structures around the metal regions in the ROI are more consistent with the reference image. These results demonstrate that LI-based conditioning, the dynamic weighting loss, and projection-consistency correction improve MAR performance through complementary structural, regional-learning, and physical-consistency constraints.

\section{Discussion and Conclusion}
\label{sec:discussion_conclusion}

A central challenge in generative MAR is to exploit the distributional prior of artifact-free CT images while preserving the patient-specific anatomy and local fidelity around metal implants. SCMA uses the linear-interpolation-corrected image as a structural condition rather than treating it as the final correction result, thereby providing a sample-specific structural anchor for the Flow Matching trajectory. This design constrains the otherwise uncertain mapping from a Gaussian initial state to an artifact-free CT image and guides the restoration process toward the anatomy of the current input.

The metal-aware dynamic weighting strategy further addresses the spatially nonuniform distribution of metal artifacts. Because severe degradation is concentrated within and around the metal regions, a spatially uniform loss may underemphasize these locally challenging areas. SCMA constructs time-dependent spatial weights from the metal mask and its distance transform, enabling velocity-field training to focus more strongly on metal-related regions while limiting unnecessary changes to reliable anatomical structures farther from the metal. During inference, projection-consistency correction constrains the generated result using the relatively reliable measurements outside the metal trace, thereby further reducing the risk of physically inconsistent hallucination-like structures.

The experimental results validate the effectiveness of these designs. SCMA consistently achieves effective artifact suppression and anatomical structure preservation across different metal sizes. It also provides higher local structural fidelity within both the metal-adjacent and spine ROIs. The ablation results show that the linear-interpolation-based structural condition is particularly important for stabilizing the conditional generation trajectory, whereas dynamic weighting and projection-consistency correction provide further improvements through localized artifact-aware learning and measurement-based physical constraints, respectively.

SCMA nevertheless has several limitations. First, the method relies on a metal mask to construct the spatial weight map and the corresponding projection-domain metal trace. Segmentation errors may therefore affect both model training and inference stability. Future work could investigate metal-aware representations that are more robust to inaccurate or uncertain masks. Second, projection-consistency correction improves physical reliability but introduces additional computational cost. More efficient approximate solvers or learned consistency operators could be explored to accelerate inference. In addition, the present method is primarily developed for two-dimensional slices and two-dimensional projection geometry. Extending SCMA to three-dimensional cone-beam or helical CT and validating it using larger, multicenter clinical datasets constitute important directions for future research.

In conclusion, we proposed SCMA, a structure-conditioned and metal-aware Flow Matching framework for CT metal artifact reduction. By integrating sample-specific structural conditioning, metal-region-aware training, and projection-consistency correction into a unified generative restoration process, SCMA effectively suppresses metal artifacts, preserves local anatomical structures, and reduces the risk of hallucination-like structures in generative MAR.

\bibliographystyle{IEEEtran}
\bibliography{cas-refs}

\end{document}